\def\year{2018}\relax
\documentclass[letterpaper]{article} 
\usepackage{aaai18}  
\usepackage{times}  
\usepackage{helvet}  
\usepackage{courier}  
\usepackage{url}  
\usepackage{graphicx}  
\usepackage{amssymb}

\usepackage{algorithm}
\usepackage[noend]{algpseudocode}
\usepackage{multirow}

\usepackage{amsmath}
\usepackage{amssymb}
\usepackage{bm}
\usepackage{balance}

\begin{document}
	%
	\title{Learning to Extract Coherent Summary via Deep Reinforcement Learning}
	\author{Yuxiang Wu$^{*}$ \\ Hong Kong University of Science and Technology\\Hong Kong\\ywubw@cse.ust.hk
	\And Baotian Hu\thanks{The two authors contribute equally to this work.}\\University of Massachusetts Medical School\\MA, USA\\Baotian.Hu@umassmed.edu}

	\maketitle
	\begin{abstract}
		 Coherence plays a critical role in producing a high-quality summary from a document. In recent years, neural extractive summarization is becoming increasingly attractive. However, most of them ignore the coherence of summaries when extracting sentences. As an effort towards extracting coherent summaries, we propose a neural coherence model to capture the cross-sentence semantic and syntactic coherence patterns. The proposed neural coherence model obviates the need for feature engineering and can be trained in an end-to-end fashion using unlabeled data. Empirical results show that the proposed neural coherence model can efficiently capture the cross-sentence coherence patterns. Using the combined output of the neural coherence model and ROUGE package as the reward, we design a reinforcement learning method to train a proposed neural extractive summarizer which is named Reinforced Neural Extractive Summarization (RNES) model. The RNES model learns to optimize coherence and informative importance of the summary simultaneously. Experimental results show that the proposed RNES outperforms existing baselines and achieves state-of-the-art performance in term of ROUGE on CNN/Daily Mail dataset. The qualitative evaluation indicates that summaries produced by RNES are more coherent and readable.
	\end{abstract}

	\section{Introduction}
	Although deep neural networks (DNN) have dominated almost every field of natural language processing, such as sentiment classification~\cite{duyutang-sentiment}, machine translation~\cite{cho-translation} and question answering~\cite{xiaoqiang}, generating high-quality summaries from long documents is still a very challenging task. Most of the recent works on abstractive summarization focus on headline generation from one paragraph~\cite{fb2015} or several sentences~\cite{lcsts} by using sequence-to-sequence architectures borrowed from neural machine translation. However, they bypass the fundamental problems in summarization, namely the representation of long documents and the generation of multiple coherent sentences. These models fail to produce readable, informative and coherent sentences when dealing with long documents. There is still a long way to go before abstractive summarization becomes practicable.
	
	In contrast, extracting sentences from documents to form summaries, also named extractive summarization, is a more practical approach, because it can guarantee the grammatical correctness of the produced summary and its semantic relevance with the corresponding document. Extractive summarization has been studied for several decades. Traditional methods mainly focus on scoring sentences using graph-based method~\cite{graph_based}, submodular functions~\cite{lin_class_2011} or integer linear programming~\cite{ilp}, which are coupled with handcrafted features. As the distributed representation shows its outstanding capability in capturing semantic and syntactic information of text~\cite{word2vec,DNV}, there is an emergence of works that use the deep neural networks to extract salient sentences~\cite{jianpeng2016,SummaRuNNer}. Although DNN-based methods can identify the important sentences from the documents, they still lack the ability to ensure coherence of the summary. They may produce summaries with sentences that are semantically independent to each other, which would cause difficulty for readers to comprehend the story as a whole. 
	
	The coherence of a summary is essential for its readability and clarity. However, to the best of our knowledge, there is no work incorporating coherence into the neural extractive model while extracting sentences. This task is challenging because it is difficult to include coherence into the objective function of supervised learning models because the coherence also depends on sentences that are eventually extracted when the inference is performed. In contrary, reinforcement learning (RL) is suitable for this case. RL algorithms aim to train an agent to maximize the reward by interacting with an environment. It is often used in settings where the objective is not differentiable with respect to the model parameters, such as works done by \cite{socher2017_summarization,rl2nmt}.
	
	In this paper, we focus on incorporating coherence into neural extractive model via reinforcement learning. We need a model that estimates coherence in the first place. During the past decades, works in coherence modeling mainly focus on topical coherence. One of the most popular methods is the entity grid model~\cite{entitygrid} which constructs a grid to represent grammatical and semantic transitions of entities between sentences. However, entity grid model depends on the named entity recognition system whose performance may become the bottleneck of entity grid model. Furthermore, entity grid models transitions of different entities separately, so it fails to capture semantic correlation between entities. Therefore, we instead use a neural coherence model which learns to estimate the coherence degree between two sentences by their distributed representation in an end-to-end fashion.

	The contribution of this paper is twofold. First, we propose a novel neural coherence model which exploits the distributed representation of sentences instead of sparse handcrafted features. The proposed neural coherence model does not rely on any entity recognition systems and can be trained from scratch in an end-to-end fashion. The neural coherence model can capture the cross-sentence local entity transitions and the discourse relations with multiple layers of convolution and max-pooling. The experimental results show that, given one sentence, the neural coherence model can effectively identify the appropriate next sentence to compose a coherent sentence pair.
	
	Second, we design a novel Reinforced Neural Extractive Summarization (RNES) model that incorporates coherence into neural extractive summarization with reinforcement learning. The output of the neural coherence model is used as immediate rewards during the training of RNES so that it learns to extract coherent summaries. ROUGE score is utilized as the final reward, and hence the proposed RNES model finds a balance between coherence and informative importance of sentences. We evaluate the proposed RNES model on CNN/Daily Mail dataset, and the results show that it achieves the state-of-the-art performance on ROUGE metrics. The qualitative evaluation indicates that the summaries produced by RNES are more informative and coherent.

	\section{Related Work}
	Our research builds on previous works in the field of neural extractive summarization, reinforcement learning, and coherence modeling.
	
	Much progress has been made beyond traditional frameworks of extractive summarization models. Most of the recent works are based on deep neural networks. For example, \cite{filippova_sentence_2015} use a recurrent neural network (RNN) to delete words from a sentence for sentence compression task. \cite{jianpeng2016} use a convolutional neural network to encode sentences, and then an RNN reads the sentence representations sequentially to encode the document. Finally, another RNN is used to label sentences sequentially, taking the encoded document representation and the previously labeled sentences into account. \cite{jianpeng2016} mainly consider the importance of sentences and the non-redundancy of the summary. \cite{SummaRuNNer} use a similar architecture to encode document, but it explicitly models sentence content, salience, novelty and position in its model for extracting sentences.
	
	
	Our work is also related to the application of reinforcement learning in document summarization. Different from classification problem whose output is a single label, the goal of extractive summarization is to make a sequence of extraction decisions. Hence, it is suitable to formulate extractive summarization as a reinforcement learning problem that tries to maximize the quality of the summaries. Although \cite{ryang_framework_2012,rioux_fear_2014,hens_s_reinforcement_2015} use value-based RL algorithms for extractive summarization, all of them are based on handcrafted features and do not consider coherence as the part of the reward. With the recent resurgence of DNN models, deep reinforcement learning has drawn considerable attention. For example, \cite{socher2017_summarization} use the ROUGE-L score as the reinforcement reward and self-critical policy gradient training algorithm to train an abstractive summarization model. \cite{ayana2016,sltrnn2016} show that directly optimizing the evaluation metrics via reinforcement learning is more effective than optimizing likelihood for the sequence generation problems. However, these works focus on abstractive summarization and neglect coherence. To our knowledge, no work has been done to apply RL to neural extractive summarization with coherence as part of the reward, and our work is the first step towards filling this gap.
	
	An essential requirement for summarization systems is the coherence of its output. Coherence is what makes multiple sentences semantically, logically and syntactically coherent~\cite{Yao2017RecentAI}.  Entity grid model proposed by \cite{entitygrid} is widely used to model the coherence of text. However, the discrete representation of entity grid suffers from the curse of dimensionality which limits its application on neural summarization. \cite{nlcm} presented a local coherence model based on a convolutional neural network that operates over the distributed representation of entity grid. Since \cite{nlcm} still rely on the entity grid features, it fails to exploit the full power of DNN in learning the hidden distributed representation of text automatically. \cite{jiweili2014} use the recurrent and recursive neural network to obtain the distributed representation of sentences and then use a pairwise ranking method to train the coherence model. This model does not need any feature engineering, but it is weak in capturing the local entity transition because of the lack of cross-sentence local interaction. Our neural coherence model can be trained from scratch in an end-to-end fashion. It can model the local entity transitions as well as the syntactic and semantic relation between sentences via different levels of cross-sentence local interaction.

	\section{Neural Extractive Summarization Model}
	
	We need to construct a neural extractive summarization (NES) model before training it with the reinforcement learning algorithm. In this section, we present the detailed architecture of the proposed NES.  
	
	
	The extractive summarization model reads the document and sequentially selects a set of sentences to compose a summary. Given a document $X=(x_1, x_2, \cdots, x_n)$ that consists of $n$ sentences, the NES model outputs a sequence of binary decisions $Y=(y_1, y_2, \cdots, y_n)$, where $n$ denotes the number of sentences in the document and $y_i \in \{0,1\}$ indicates whether sentence $x_i$ is selected. Then the extracted summary is a sequence of $l$ sentences denoted as
	\[G=\text{extract}(X, Y)=(x_{q_1}, \cdots, x_{q_l}),\]
	where $1\leq q_1 < \cdots < q_l \leq n$, and $y_{q_i} = 1$ for $i=1,\cdots,l$.
	
	The proposed NES uses a hierarchical deep neural network to encode the document. At the word-level, convolutional neural network (CNN) is used to extract features of the words and their context. Let $x_t=(w_1, w_2, \cdots, w_m)$ denotes the $t$-th sentence with $m$ words, and $v$ denotes the size of word embedding. Then the sentence could be represented by a matrix $\mathbf{M} \in \mathbb{R}^{m \times v}$. Multiple convolution kernels with different kernel size are used to extract features of word $w_i$:
	$$\mathbf{f}_i^j = \mathbf{M}_{i:i+k_j-1} \mathbf{W}_j + \mathbf{b}_j ,$$
	where $\mathbf{W}_j, \mathbf{b}_j, k_j$ are the kernel weight matrix, the bias and the kernel size of the $j$-th convolution kernel respectively. The word $w_i$ is represented by concatenating the feature maps $\mathbf{f}_{w_i}=[\mathbf{f}_i^1 ; \mathbf{f}_i^2 ; \cdots]$. The sentence $x_t$ is represented by the mean of all its word features
	$$ \mathbf{x}_t = \frac{1}{m} \sum_{i=1}^m \mathbf{f}_{w_i} .$$
	
	At the sentence-level, we use a bi-directional gated recurrent unit (Bi-GRU) to model the context of sentences. Gated recurrent unit is a variant of recurrent neural network proposed by \cite{chung2014empirical}. It has two gates, an update gate $\mathbf{z}_t$ and a reset gate $\mathbf{r}_t$. The hidden state $\mathbf{h}_t$ at time step $t$ could be computed with following equations:
	\[ \mathbf{z}_t = \sigma(\mathbf{W}_{z} \mathbf{x}_t + \mathbf{V}_{z} \mathbf{h}_{t-1}  + \mathbf{b}_{z}) , \]
	\[ \mathbf{r}_t = \sigma(\mathbf{W}_{r} \mathbf{x}_t + \mathbf{V}_{r} \mathbf{h}_{t-1}  + \mathbf{b}_{r}) , \]
	\[ \hat{\mathbf{h}}_t = \tanh(\mathbf{W}_{h} \mathbf{x}_t + \mathbf{V}_{h} (\mathbf{r}_{t} \odot \mathbf{h}_{t-1} ) + \mathbf{b}_{h} ) ,\]
	\[ \mathbf{h}_t = (1 - \mathbf{z}_t) \odot \hat{\mathbf{h}}_{t} +  \mathbf{z}_t \odot \mathbf{h}_{t-1} ,\]
	where $\odot$ represents element-wise product, $\mathbf{W}_{*}$'s, $\mathbf{V}_{*}$'s and $\mathbf{b}_{*}$'s are parameters of GRU. 
	
	Using Bi-GRU, the representation of the $t$-th sentence $\mathbf{x}_t$ is transformed to a forward hidden state $\overrightarrow{\mathbf{h}}_t$ and a backward hidden state $\overleftarrow{\mathbf{h}}_t$. Both states are concatenated to form the contextual representation of the $t$-th sentence $\overleftrightarrow{\mathbf{h}}_t = [\overrightarrow{\mathbf{h}}_t ; \overleftarrow{\mathbf{h}}_t]$. The entire document is represented as $\mathbf{d}$ by a nonlinear transformation of the mean over all sentence representations:
	\[ \mathbf{d} = \tanh( \mathbf{W}_d (\frac{1}{n} \sum_{t=1}^{n} \overleftrightarrow{\mathbf{h}}_t ) + \mathbf{b}_d ) ,\]
	where $\mathbf{W}_d$ and $\mathbf{b}_d$ are parameters of the transformation. 
	
	The probability of extraction decisions $Y$ conditioned on document $X$ could be factorized as $\Pr(Y|X) = \prod_{t=1}^{n} \Pr(y_t | X, y_{1:t-1})$.
	The probability of extracting the $t$-th sentence is computed as
	\begin{equation} \label{eq:mlp}
		 \Pr(y_t=1|X, y_{1:t-1}) = \text{MLP}(\overleftrightarrow{\mathbf{h}}_t, \mathbf{g}_{t-1}, \mathbf{d} ) ,
	\end{equation}
	where $\mathbf{g}_{t-1}$ represents all sentences extracted before time $t$. $\text{MLP}(\cdot)$ means a multilayer perceptron that outputs a probability
	\begin{align*}
	\text{MLP}(\overleftrightarrow{\mathbf{h}}_t, \mathbf{g}_{t-1}, \mathbf{d} ) =  \sigma(  & \mathbf{W}_2 \tanh(\mathbf{W}_1 [\overleftrightarrow{\mathbf{h}}_t, \mathbf{g}_{t-1}, \mathbf{d}] \\
	& + \mathbf{b}_1) + \mathbf{b}_2 ),
	\end{align*}
	where $\mathbf{W}_{1},\mathbf{W}_{2}, \mathbf{b}_{1}$ and $\mathbf{b}_{2}$ are parameters of the MLP and $\sigma(\cdot)$ is the sigmoid function. 
	
	Since NES is trained with supervised learning, ground truth extraction labels $(\mathring{y}_1, \cdots, \mathring{y}_n)$ are available during training. Then the representation of sentences selected before or at time $t$ is
	\begin{equation*}
	\mathbf{g}_t =  \mathbf{g}_{t-1} + \mathring{y}_t \tanh (\mathbf{W}_{g} \overleftrightarrow{\mathbf{h}}_t) .
	\end{equation*}
	
	The NES model is pretrained by minimizing the negative log-likelihood of the ground truth extraction labels
	\begin{align*}
	L_{\text{pretrain}}(\Theta) = - \sum_{i=1}^{N} &\sum_{t=1}^{N_i} \big[ \mathring{y}_t^i \log \Pr(y_t^i=1|X_i, \mathring{y}_{1:t-1}^i)\\ 
	 + & (1 - \mathring{y}_t^i) \log \Pr(y_t^i=0|X_i, \mathring{y}_{1:t-1}^i)\big] .
	\end{align*}
	
	\section{Reinforced Neural Extractive Summarization Model}
	\label{sec:rl}
    After the NES model is pretrained with supervised learning, we further train it with reinforcement learning to extract coherent and informative summaries by maximizing coherence and ROUGE scores. In this section, we first introduce the REINFORCE algorithm and then describe the proposed neural coherence model and the ROUGE score reward. The overall training algorithm is illustrated in Algorithm \ref{algo_rnes}.

	\subsection{Reinforcement Learning} 

	The problem of extractive summarization could be formulated as a reinforcement learning problem. The RNES model can be considered as an \emph{agent} that extracts sentences sequentially from the document. At each time step $t$, the agent is in \emph{state} $s_t = (X, y_{1:t-1})$ which includes the document and the previous selections. Agent would take an \emph{action} $y_t \in \{0,1\}$ that decides whether sentence $x_t$ is extracted or not. After the agent takes the action $y_t$, it may receive an immediate reward $r_{t}$ that shows how good the action is. The reward could also be delayed. When the agent finishes extracting sentences from the document, it will receive a final reward $r_{-1}$ that indicates the performance of the entire action sequence $(y_1, y_2, \cdots, y_n)$.
	
	We use the REINFORCE algorithm to train our RNES model. It is a kind of policy gradient method proposed by  \cite{williams_simple_1992}, and it maximizes the performance of the agent by updating its policy parameters. The policy is defined as the probability of taking an action at time $t$ given a state, which is parameterized by $\Theta$:
	\begin{align*}
	\pi(a|s_t,\Theta) &\overset{\text{def}}{=} \Pr(y_t=a|s_t,\Theta) \\
	&\overset{\text{def}}{=} \Pr(y_t=a|X, y_{1:t-1}, \Theta) .
	\end{align*}
	In our case, $\Theta$ represents all the parameters in the RNES model. We use a shorthand $\pi_{\Theta}$ to denote the policy $\pi$ parameterized by $\Theta$. By applying Equation \ref{eq:mlp}, we have
	\begin{equation*}
	\pi_{\Theta}(a=1|s_t) = \text{MLP}_{\Theta}(\overleftrightarrow{\mathbf{h}}_t, \mathbf{g}_{t-1}, \mathbf{d} ) .
	\end{equation*}
	
	Let $s_0=X$ represents the initial state when no action is taken yet, and $v_{\pi_{\Theta}}(s_0)$ be the \emph{value} function that represents the expected \emph{return} starting with state $s_0$ by following policy $\pi_{\Theta}$. Return at time $t$ is defined as $R_t=\sum_{i=t}^{\infty} \gamma^{i-t} r_i$, where $\gamma$ is the discount factor. The objective of REINFORCE is defined as maximizing the value of initial state $v_{\pi_{\Theta}}(s_0)$, or minimizing its negative $L_{RF}(\Theta) = - v_{\pi_{\Theta}}(s_0)$. Therefore, the parameters should be updated by the gradient of $L_{RF}(\Theta)$ with respect to parameters $\Theta$:
	\begin{align*}
	\nabla L_{RF}(&\Theta) = - \nabla v_{\pi_{\Theta}}(s_0) \\ 
	=& - \sum_{t=1}^{n} \gamma^t \Pr(s_t | s_0, \pi_{\Theta}) \sum_{a} q_{\pi_{\Theta}}(s_t,a) \nabla \pi_{\Theta}(a|s_t) ,
	\end{align*} 
	where $q_{\pi_{\Theta}}(s,a)$ is the \emph{action-value} function that represents the expected return after taking action $a$ at state $s$ with policy $\pi_{\Theta}$. 
	
	Since the state space is too large, it is infeasible to compute the exact value of the gradient. We use Monte Carlo sampling to approximate the gradient:
    \begin{equation} \label{eq:MCRF}
	\nabla L_{RF}(\Theta) = - \mathbb{E}_{\tilde{y}_t, \tilde{s}_t \sim \pi_{\Theta}} \Big[ \gamma^t \tilde{R}_t \nabla \log \pi_{\Theta}(\tilde{y}_t|\tilde{s}_t) \Big] ,
	\end{equation}
	where $\tilde{s}_t$ and $\tilde{y}_t$ are randomly sampled from $\pi_{\Theta}$, $\tilde{R}_t$ is the actual return received since $\tilde{s}_t$ and $\tilde{y}_t$. A detailed proof of Equation \ref{eq:MCRF} could be found in \cite{sutton_reinforcement_nodate} and is omitted for brevity here. The parameters $\Theta$ are updated as follows:
	\begin{equation} \label{eq:update}
	\Theta \leftarrow \Theta + \gamma^t \tilde{R}_t \nabla \log \pi_{\Theta}(\tilde{y}_t|\tilde{s}_t) .
	\end{equation}
	We use $\gamma=1$ for simplicity in this work.
	
	The definition of reward is crucial for reinforcement learning because it determines the optimization direction. To ensure that the RNES model extracts coherent and informative summaries, the reward includes both coherence score and ROUGE score, which will be introduced later in this paper. Given a sequence of sampled actions $\tilde{Y} = (\tilde{y}_1, \cdots, \tilde{y}_n)$, the corresponding coherence scores are exploited as immediate rewards $\tilde{r}_t$ and the ROUGE score as the final reward $\tilde{r}_{-1}$. Therefore, our algorithm is indeed maximizing a weighted sum of coherence and ROUGE score:
	\begin{align}
	v_{\pi}(s_0) &\overset{\text{def}}{=} \mathbb{E}_{\pi_{\Theta}}[R_0 |s_0] = \mathbb{E}_{\tilde{y}_t, \tilde{s}_t \sim \pi_{\Theta}} [ \tilde{r}_{-1} + \lambda \sum_{t=1}^{n} \tilde{r}_t | s_0] \nonumber \\
	&= \mathbb{E}_{\pi_{\Theta}} [ \text{ROUGE}(\tilde{G}) + \lambda \text{Coherence}(\tilde{G}) | X ] \label{eq:reward_sum} 
	\end{align}
	where $\tilde{G} =\text{extract} (X, \tilde{Y})$ is the sampled extractive summary and $\lambda$ is the coefficient that balances the two rewards. $\text{Coherence}(\tilde{G})$ is the sum of coherence scores of $\tilde{G}$:
	\[\text{Coherence}(\tilde{G})= \sum_{(\tilde{S}_A, \tilde{S}_B) \in \text{adj}(\tilde{G})} \text{Coh}(\tilde{S}_A, \tilde{S}_B) ,\]
 	where $\text{adj}(\tilde{G})$ is the set of adjacent sentences in $\tilde{G}$. The function $\text{Coh}(\cdot,\cdot)$ is defined by the neural coherence model in Equation \ref{eq:coh}, which will be introduced in the next subsection. Algorithm~\ref{algo_rnes} shows the overall REINFORCE algorithm to train our proposed RNES model.  
	
	\begin{algorithm}[t]
		\small
		\begin{algorithmic}[1]
			\State $\Psi \leftarrow$ train the neural coherence model.
			\State $\Theta \leftarrow$ pretrain the neural sentence extractor with supervised learning.
			
			\Loop
			
			\State $X, H \leftarrow$ sample a document-summary pair from corpus
			\State $\tilde{s}_0 \leftarrow X$
			\State Sample an episode $\tilde{s}_1, \tilde{y}_1, \cdots, \tilde{s}_n, \tilde{y}_n$ following $\pi_{\Theta}$
			
			\State $previous \leftarrow \chi$ (a placeholder for empty start sentence)
			\For  {each step $t=1 \dots n$}
			\If {$\tilde{y}_t = 1$}
			\State $\tilde{r}_t \leftarrow \text{Coh}_{\Psi}(previous, x_t)$
			\State $previous \leftarrow x_t$
			\Else
			\State $\tilde{r}_t \leftarrow 0 $
			\EndIf
			\EndFor

			\State $\tilde{G} \leftarrow$ extract($X, (\tilde{y}_1, \cdots, \tilde{y}_n)$)
			\State $\tilde{r}_{-1} = \text{ROUGE}(\tilde{G}, H)$
			
			\For  {each step $t=1 \dots n$}
			\State $\tilde{R}_t \leftarrow \lambda \sum_{i=t}^{n} \tilde{r}_i + \tilde{r}_{-1}$
			\State $\Theta \leftarrow \Theta + \alpha \tilde{R}_t \nabla \log \pi_{\Theta}(\tilde{y}_t|\tilde{s}_t)$
			\EndFor
			
			\EndLoop
			
		\end{algorithmic}
		\caption{Overall training algorithm of RNES model. $\alpha$ is the learning rate, $\chi$ is a placeholder sentence for bootstrapping the coherence score of the first extracted sentence.}
		\label{algo_rnes}
	\end{algorithm}

	\subsection{Neural Coherence Reward}
	\label{ncr}
\begin{figure}
		\includegraphics[width=\columnwidth]{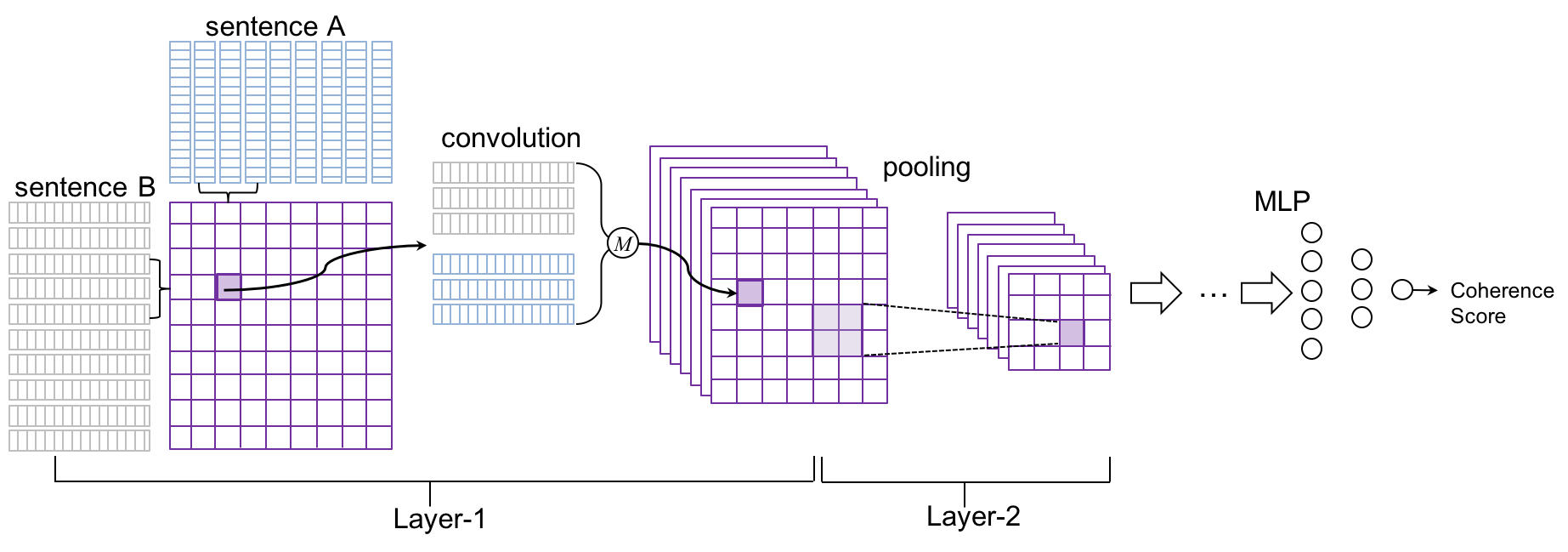}
		\caption{Illustration of neural coherence model which is built upon ARC-II proposed by~\cite{NIPS2014_hu}.}
		\label{coherence}
	\end{figure}

	We propose a neural coherence model to compute the cross sentence coherence as part of the reward of RNES model. This model is built on the ARC-II proposed by \cite{NIPS2014_hu} for sentence matching. This neural coherence model has some advantages over traditional entity grid models. Our neural coherence model requires no feature engineering and could be trained in an end-to-end fashion. Besides, it uses the distributed text representation which can capture the syntactic and semantic coherence patterns by cross-sentence interaction. The architecture of the neural coherence model is shown in Figure~\ref{coherence}. 

	Given two sentences $S_A$ and $S_B$, in layer 1, it uses sliding windows on both sentences to model all the possible local coherence transition of the two sentences. For segment $i$ on $S_A$ and segment $j$ on $S_B$, the local coherence transition is computed as
	\begin{equation*}
	\mathbf{z}^{(1)}_{i,j}  = \text{ReLU}(\mathbf{W}^{(1)} \hat{\mathbf{z}}^{(0)}_{i,j} + \mathbf{b}^{(1)}),
	\end{equation*}
	where $\mathbf{W}^1$ is the weight parameters for first layer and $\mathbf{b}^1$ is the bias. ReLU is the nonlinear function proposed by~\cite{relu}. $\hat{\mathbf{z}}^{(0)}_{i,j} \in \mathbb{R}^{2k_1 \times D_{e}}$ is obtained by concatenating the embeddings of words in $S_A$ and $S_B$ sequentially:
	\begin{equation*}
	\hat{\mathbf{z}}^{(0)}_{i,j} =  [e(a_i); ... ; e(a_{i+k_1-1}); e(b_j);...; e(b_{j+k_1-1})],
	\end{equation*}
	where $a_i$ is the $i$-th word of $S_A$, $b_j$ is the $j$-th word of $S_B$ and $e(\cdot)$ is the embedding lookup function which outputs a $D_e$-dimensional word embedding.
	
	Layer 2 takes the output of layer 1 and performs a max-pooling in each dimension on non-overlapping $2\times2$ windows.
	\begin{equation*}
	\mathbf{z}_{i,j}^{(2)} = \max(\mathbf{z}_{2i-1,2j-1}^{(1)}, \mathbf{z}_{2i-1,2j}^{(1)},\mathbf{z}_{2i,2j-1}^{(1)},\mathbf{z}_{2i,2j}^{(1)}). 
	\end{equation*}
	
	Following layer 2, there are more convolution and  max-pooling layers, analogous to that of convolutional architecture for image input~\cite{cnn}. Finally, we obtain the fixed length vector $\mathbf{h}$ and it is fed into a nonlinear transformation with activation function $\tanh$ to compute coherence score of the two sentences:
	\begin{equation}\label{eq:coh}
	\text{Coh}(S_A,S_B) = \tanh(\mathbf{W}_c  \mathbf{h}+ \mathbf{b}_c), 
	\end{equation}
	where $\mathbf{W}_c$ is the weight parameters and $\mathbf{b}_c$ is the bias. Hence, the coherence model will output a coherence score  $\text{Coh}(S_A, S_B) \in (-1,1)$ for any sentence pairs $(S_A, S_B)$.
	
	From the first layer, the neural coherence model can capture the local coherence of two sentences.  And it can also obtain higher level coherence representation of $S_A$ and $S_B$ with more convolution and max-pooling layers.
	
	For the training of the neural coherence model, we use a pair-wise training strategy with a large margin objective. 
	Suppose we are given the following triples $(S_A, {S_B}^{+},{S_B}^{-})$, we adopt the ranking-based loss as objective:
	\begin{align*}
	\begin{split}
	L_\Theta({S_A}, {S_B}^+, {S_B}^-) = 
	\max(0, 1 &+\text{Coh}({S_A, {S_B}^-})\\
	&-\text{Coh}({S_A}, {S_B}^+)).
	\end{split}
	\end{align*}
	The model is trained by minimizing the above objective, to encourage the model to assign higher coherence score to coherent sentence pair $(S_A, {S_B}^+)$ than incoherent pair $({S_A}, {S_B}^-)$.
	

	\subsection{ROUGE Score Reward}
	\label{rsr}
	ROUGE score is used as the final reward to ensure that RNES model extracts reasonably informative sentences. Given a sequence of sampled decisions $(\tilde{y}_1, \cdots, \tilde{y}_n)$, we can get the sequence of extracted sentences $\tilde{G}$. Since the dataset comes with news highlights written by human editors, these manual highlights $H$ is treated as the reference summary. Then the ROUGE score between the system summary $\tilde{G}$ and the reference $H$  could be computed and used as the final reward for the entire sampled decisions:
	\[ \tilde{r}_{-1}(\tilde{y}_1, \cdots, \tilde{y}_n) = \text{ROUGE}(\tilde{G}, H). \]

	Multiple variants of ROUGE score are proposed by \cite{rouge}. Among them, ROUGE-1 (R-1), ROUGE-2 (R-2) and ROUGE-L (R-L) are the most commonly used ones. ROUGE-$n$ (R-$n$) recall between an extracted summary and a reference summary can be computed as follows:
	\[ \text{R-}n = \frac{\sum_{s\in\text{reference summary}} \sum_{\text{gram}_n \in s} \text{Count}_{\text{match}} (\text{gram}_n) }{\sum_{s\in\text{reference summary}} \sum_{\text{gram}_n \in s} \text{Count} (\text{gram}_n)} , \]
	where $n$ stands for the length of n-gram, $\text{Count}_{\text{match}} (\text{gram}_n)$ is the maximum number of n-grams co-occurring in both the extracted summary and the reference. Similarly we could compute the R-$n$ precision and F1. R-1 and R-2 are special cases of R-$n$ in which $n=1$ or $n=2$. R-L is instead computed based on the length of longest common subsequence between the system summary and the reference. Since using only one variant of ROUGE as reward for training RNES may not increase its performance on other ROUGE variants, we use a combination of ROUGE variants as reward:
	\begin{align*}
	\text{ROUGE}(G, H) = & w_1 \text{R-1}(G, H) + w_2 \text{R-2}(G, H) \\
	& + w_l \text{R-L}(G, H),
	\end{align*}
	where weights $w_1$, $w_2$ and $w_l$ are hyperparameters. We use $w_1=0.4, w_2=1.0, w_l=0.5$ in our experiments to ensure balanced enhancement.

	The overall training algorithm is illustrated in the Algorithm \ref{algo_rnes}. Since REINFORCE algorithm converges very slowly, we pretrain the RNES model with supervised learning. The neural coherence model is also trained and then fixed for the coherence scoring. During the REINFORCE training, a sequence of actions and states is sampled according to the policy. Then the coherence model and the ROUGE package are used for computing the rewards. The parameters of RNES model $\Theta$ is then updated according to Equation \ref{eq:update}.

	\section{Experiments and Results}

	We use the CNN/Daily Mail dataset originally introduced by \cite{hermann_teaching_2015} to evaluate our model. This dataset contains news documents and their corresponding highlights crawled from CNN and Daily Mail website, and it is commonly used in extractive summarization~\cite{jianpeng2016,SummaRuNNer} and abstractive summarization~\cite{nallapati_ramesh_abstractive_2016,see_get_2017}. We used the scripts provided by \cite{hermann_teaching_2015} to download the dataset. It contains 287,226 documents for training, 13,368 documents for validation and 11,490 documents for test. Since the dataset only contains manual summaries and does not have extractive labels, a greedy algorithm similar to the one presented by \cite{SummaRuNNer} is used to generate extraction labels for supervised training.
	 
	\subsection{Results of the Neural Coherence Model}
	The coherence model needs to be trained before it is used to produce coherence score as the reward in the REINFORCE algorithm. In our experiments,  we use 64-dimensional word embeddings which are randomly initialized and finetuned in the process of supervised training. The sizes of all its convolutional kernels are set to 3. The first convolution layer has 128 filters. The second and third convolution layers contain 256 and 512 filters respectively. Each convolution layer is followed by a max-pooling layer performed on the sliding non-overlapping $2\times2$ windows. The final two fully-connected layers have 512 and 256 hidden units respectively. The maximum sentence length is 50. Sentences longer than the limit would be truncated, and those that are shorter than this length would be padded with zeros. The coherence model is trained with stochastic gradient descent (SGD) with batch size 64 and learning rate 0.1. 
	
	The training triplets are sampled from the CNN/Daily Mail dataset. The $S_A$ and ${S_B}^{+}$ are adjacent sentences sampled from the documents, and ${S_B}^{-}$ is a sentence randomly sampled such that ${S_B}^{-} \neq {S_B}^{+}$. To make the task more difficult so that the model finds more fine grained coherence patterns, ${S_B}^{-}$ is sampled from the same document as $(S_A, {S_B}^{+})$ and it is less than nine sentences away from ${S_B}^{+}$.
	
	The model is tested on around twenty three thousand positive pairs sampled from the test set, each accompanied with one negative sample. If the model gives a higher score to the positive sample than the negative sample, it is considered correct. The accuracy is 71.3\%, versus 50\% accuracy for random guess, which indicates that the neural coherence model can capture the cross-sentence coherence.
	
	We also conducted empirical studies on some example outputs of our proposed neural coherence model. Table \ref{tab:coherence_examples} shows some examples of coherence scoring. The first example shows that the model can exploit co-reference for coherence modeling. The model can also capture the semantic coherence between two sentences, such as semantically related words such as ``photographer'' and ``shoot'' (example 1), ``survey'' and ``answers'' (example 3). Furthermore, the coherence model can discover syntactic patterns such as ``As a result \dots'' and ``According to \dots'', which represents the syntactic coherence across sentences. The third example also shows that there is much noise in our training data. After closer inspection, we found that the ${S_B}^{-}$ rather than ${S_B}^{+}$, is the right sentence following ${S_A}$. ${S_B}^{+}$ in example 3 is an image caption embedded in the article, which is not filtered out during the data preprocessing phase. However, thanks to the training on the large-scale text corpus, the neural coherence model is robust enough to score the right sentence much higher. These examples show that the neural coherence model indeed captures the semantic and syntactic coherence patterns across sentences.

      \begin{table}[htb]
		\centering
		\caption{Example outputs of neural coherence model.}
		\label{tab:coherence_examples}
		\begin{tabular}{|p{65mm}|p{10mm}|}
			\hline
			 \centering Sentences &  Score \\\hline
			$S_A$: \small{Terry's career as a \textbf{photographer} came after he failed to make it as a punk rock musician.} & \\
			${S_B}^{+}$: \small{\textbf{He} got his first big break in 1994 with a \textbf{shoot} for Vibe magazine.} & \textbf{0.9885} \\
			${S_B}^{-}$: \small{The photographer has also directly music videos in his time.} & 0.5198 \\
			 \hline
			 $S_A$: \small{These days we are increasingly using outdoor space for the occasional barbecue or to relax in a hot tub rather than for tending \textbf{flowers}, according to researchers.} & \\
			 ${S_B}^{+}$: \small{\textbf{As a result}, only a handful of traditional \textbf{flowers} still grow in English country gardens, with the average one usually containing a mere four species - \textbf{daffodils}, \textbf{crocuses}, \textbf{roses} and \textbf{tulips}.} & \textbf{0.8934} \\
			 ${S_B}^{-}$: \small{Sir Roy Strong, the landscape designer and former director of the Victoria and Albert Museum, told the Sunday Times: `British people used to take pride in having neat gardens with lots of flowers.'} & -0.0067 \\
			\hline
			$S_A$: \small{The same \textbf{survey} recently showed that university pupils in Britain have an average of 8.2 sexual partners by the time they reach the middle of their higher education.} & \\
			${S_B}^{+}$: \small{A new survey of university students has revealed that they have had an average of 8.2 sexual partners (picture posed by models)} & 0.0021 \\
			${S_B}^{-}$: \small{\textbf{According to the answers they received}, 22 per cent of students didn't lose their virginity until they were 18 years old, with the second most popular age to have sex for the first time being 16. } & \textbf{0.9999} \\
			\hline
		\end{tabular}
	\end{table}
	
	\subsection{Results of the Reinforced Neural Extractive Summarization Model (RNES)}
	For the NES/RNES model, we use 128-dimensional word embeddings and the vocabulary size is 150,000. The convolution kernels have size 3, 5, 7 with 128, 256, 256 filters respectively. We set the hidden state size of sentence-level GRU to 256, and the document representation size to 512. The MLP has two layers, with 512 and 256 hidden units respectively. We fix the maximum sentence length to 50 and the maximum number of sentences in a document to 80. Sentences or documents that are longer than the maximum length are truncated to fit the length requirement.
	
	The model is trained with stochastic gradient descent (SGD) with batch size 64. When doing supervised training of the NES, ground truth extraction labels are used for computing the cross-entropy loss. The labels are generated from the dataset by greedily selects sentences to maximize its ROUGE similarity compared to manual highlights.
	
	During the training of RNES using reinforcement learning, both the neural coherence model and ROUGE scorer are used to compute the reward. As shown in Equation \ref{eq:reward_sum}, the hyperparameter $\lambda$ is used to balance between the two objectives. In our experiments, we explored $\lambda=1.0, 0.1, 0.01, 0.005$. It is found that when $\lambda=1.0$ or $0.1$, the model favors coherence so much that ROUGE degrades rapidly and the model eventually converges to a policy that selects consecutive sentences that are not informative. However, when $\lambda = 0.005$, the ROUGE objective overpowers coherence, and the coherence rewards drop to approximately zero. We found that $\lambda=0.01$ is a good trade-off such that both rewards increase and eventually converge.
	
	At test time our model produces summaries by beam search with beam size 10. To compare with previous works, we adopt the same evaluation metrics as in \cite{SummaRuNNer}. We use full-length F1 of ROUGE-1, ROUGE-2, and ROUGE-L to evaluate our model. Table \ref{tab:rouge_cnn_dm} shows the performance comparison between our models and baselines. Our RNES models (with or without coherence as the reward) outperform current state-of-the-art models and NES by a large margin. The result indicates that the summaries extracted by RNES are of higher quality than summaries produced by previous works.

	\begin{table}[ht]
		\centering
		\caption{Performance comparison on CNN/Daily Mail test set, evaluated with full-length F1 ROUGE scores (\%). All scores of RNES are statistically significant using 95\% confidence interval with respect to previous best models.}
		\label{tab:rouge_cnn_dm}
		\begin{tabular}{|p{32mm}|p{12mm}|p{12mm}|p{11mm}|}
			\hline
			Model&R-1&R-2&R-L\\\hline
			Lead-3&39.2&15.7&35.5\\
			\cite{nallapati_ramesh_abstractive_2016}&35.4&13.3&32.6\\
			(Nallapati et al. 2017) &39.6&16.2&35.3\\
			(See et al. 2017)&39.53&17.28&35.38\\
			NES & 37.75 & 17.04 & 33.92\\\hline
			RNES w/o coherence&\textbf{41.25}&\textbf{18.87}&\textbf{37.75}\\
			RNES w/ coherence&40.95&18.63&37.41\\\hline
		\end{tabular}
	\end{table}

	Though RNES with the coherence reward achieves higher ROUGE scores than baselines, there is a small gap between its score and that of RNES trained without coherence model. This is because that the coherence objective and ROUGE score do not always agree with each other. Since ROUGE is simply computed based on n-grams or longest common subsequence, it is ignorant of the coherence between sentences. Therefore, enhancing coherence may lead to a drop of ROUGE. However, the 95\% confidence intervals of the two RNES models overlap heavily, indicating that their difference in ROUGE is insignificant.

	\begin{table}[ht]
	\centering
	\caption{Comparison of human evaluation in terms of informativeness(Inf), coherence(Coh) and overall ranking. Lower is better.}
	\label{tab:human_eval}
	
	\begin{tabular}{|p{31mm}|p{12mm}|p{12mm}|p{12mm}|}
		\hline
		Model&Inf&Coh&Overall\\\hline
		RNES w/o coherence&1.183&1.325&1.492 \\
		RNES w/ coherence& \textbf{1.125} & \textbf{1.092} & \textbf{1.209} \\\hline
	\end{tabular}
	\end{table}

	We also conduct a qualitative evaluation to find out whether the introduction of coherence reward improves the coherence of the output summaries. We randomly sample 50 documents from the test set and ask three volunteers to evaluate the summaries extracted by RNES trained with or without coherence as the reward. They are asked to compare and rank the outputs of two models regarding three aspects: informativeness, coherence and overall quality. The better one will be given rank 1, while the other will be given rank 2 if it is worse. In some cases, if the two outputs are identical or have the same quality, the ranks could be tied, i.e., both of them are given rank 1. Table \ref{tab:human_eval} shows the results of human evaluation. RNES model trained with coherence reward is better than RNES model without coherence reward in all three aspects, especially in the coherence. The result indicates that the introduction of coherence effectively improves the coherence of extracted summaries, as well as the overall quality. It is surprising that summaries produced by RNES with coherence are also more informative than RNES without coherence, indicating that ROUGE might not be the gold standard to evaluate informativeness as well. 
	
	Table \ref{tab:summary_examples} shows a pair of summary produced by RNES with or without coherence. The summary produced by RNES without coherence starts with pronoun `That' which is referring to a previously mentioned fact, and hence it may lead to confusion. In contrast, the output of RNES trained with coherence reward includes the sentence ``The earthquake disaster \dots'' before referring to this fact in the second sentence, and therefore is more coherent and readable. This is because the coherence model gives a higher score to the second sentence if it can form a coherent sentence pair with the first sentence. In REINFORCE training, if the second sentence receives a high coherence score, the action of extracting the first sentence before the second one will be strengthened. This example shows that coherence model is indeed effective in changing the behavior of RNES towards extracting summaries that are more coherent.
		
	\begin{table}[ht]
		\centering
		\caption{Examples of extracted summary.}
		\label{tab:summary_examples}
		
		\begin{tabular}{|p{80mm}|}
			\hline
			\small{\textit{Reference}: Peter Spinks from the Sydney Morning Herald reported on Amasia. Within 200 million years, he said the new supercontinent will form. One researcher recently travelled to Nepal to gather further information. He spotted that India, Eurasia and other plates are slowly moving together.} \\\hline
			
			\small{\textit{RNES w/o coherence}: That's according to one researcher who travelled to the country to study how the Indian and Eurasian plates are moving together. And using new techniques, researchers can now start examining the changes due to take place over the next tens of millions of years like never before. Earth's continents are slowly moving together, and in 50 to 200 million years they are expected to form a new supercontinent called Amasia. In 2012 a study suggested this may be centered on the North Pole. The idea that Earth is set to form a new supercontinent-dubbed Amasia - is not new.}\\\hline

			\small{\textit{RNES w/ coherence}: \textbf{The earthquake disaster in Nepal has highlighted how Earth's land masses are already in the process of forming a new supercontinent. That}'s according to one researcher who travelled to the country to study how the Indian and Eurasian plates are moving together. And using new techniques, researchers can now start examining the changes due to take place over the next tens of millions of years like never before. Earth's continents are slowly moving together, and in 50 to 200 million years they are expected to form a new supercontinent called Amasia.} \\
			\hline
		\end{tabular}
	\end{table}

	\section{Conclusion}	
	In this paper, we proposed a Reinforced Neural Extractive Summarization model to extract a coherent and informative summary from a single document. Empirical results show that the proposed RNES model can balance between the cross-sentence coherence and importance of the sentences effectively, and achieve state-of-the-art performance on the benchmark dataset. For future work, we will focus on improving the performance of our neural coherence model and introducing human knowledge into the RNES.    
	
	\section*{Acknowledgments}
	This work is supported by grants from WeChat-HKUST Joint Lab on Artificial Intelligence Technology (WHAT Lab). Baotian Hu acknowledges partial support from the University of Massachusetts Medical School.
	
	\bibliographystyle{aaai}
	\bibliography{ref}
	
\end{document}